\documentclass{article}
\usepackage{arxiv}

\usepackage{amsmath}
\usepackage{amsthm}
\usepackage{amssymb}
\usepackage{graphicx}
\usepackage{booktabs}
\title{Balancing Complexity and Informativeness in LLM-Based Clustering: Finding the Goldilocks Zone}

\author{
 Justin K. Miller \\
  School of Physics\\
  University of Sydney\\
  Camperdown, NSW 2006 \\
  \texttt{justin.k.miller@sydney.edu.au} \\
   \And
 Tristram J. Alexander \\
  School of Physics\\
  University of Sydney\\
  Camperdown, NSW 2006 \\
  \texttt{tristram.alexander@sydney.edu.au} \\
}

\begin{document}
\maketitle


\begin{abstract} 
The challenge of clustering short text data lies in balancing informativeness with interpretability. Traditional evaluation metrics often overlook this trade-off. Inspired by linguistic principles of  communicative efficiency, this paper investigates the optimal number of clusters by quantifying the trade-off between informativeness and cognitive simplicity. We use large language models (LLMs) to generate cluster names and evaluate their effectiveness through semantic density, information theory, and clustering accuracy.

Our results show that Gaussian Mixture Model (GMM) clustering  on embeddings generated by a LLM, increases semantic density compared to random assignment, effectively grouping similar bios. However, as clusters increase, interpretability declines, as measured by a generative LLM’s ability to correctly assign bios based on cluster names. A logistic regression analysis confirms that classification accuracy depends on the semantic similarity between bios and their assigned cluster names, as well as their distinction from alternatives.

These findings reveal a “Goldilocks zone” where clusters remain distinct yet interpretable. We identify an optimal range of 16–22 clusters, paralleling linguistic efficiency in lexical categorization. These insights inform both theoretical models and practical applications, guiding future research toward optimising cluster interpretability and usefulness.

\end{abstract}

\section{Introduction}
%
The ``Goldilocks zone" in language is a powerful analogy for understanding the optimal selection of clusters ($K$) when performing text clustering. 

This zone represents an equilibrium between two competing forces: simplicity, defined as having fewer, broader categories, and informativeness, the capacity to express precise and meaningful distinctions \cite{aceves2024human, brochhagenCoevolutionLexicalMeaning2018}. In linguistics, there is evidence that simplicity—the ease of learning fewer categories—is often initially favoured over informativeness—the ability to precisely express nuanced distinctions \cite{carr2020simplicity}. However, languages continually evolve under pressures to balance these two factors, achieving a state that is neither overly simplistic nor excessively detailed. This dynamic interaction between lexical simplicity and the ability to accurately describe something reflects evolutionary pressures for communicative efficiency and cognitive ease of acquisition \cite{brochhagenCoevolutionLexicalMeaning2018}.

This equilibrium is quantitatively grounded in linguistic theories of efficiency, where semantic systems, such as colour naming and numerical systems, often evolve toward Pareto-optimal configurations \cite{regier2007color, zaslavsky2018efficient, gibson2017color}. Pareto frontiers represent optimal states where any attempt to simplify further dramatically decreases communicative clarity, while adding complexity provides only marginal communicative benefit. For instance, colour naming across languages compresses infinite perceptual possibilities into a limited lexicon, closely approximating solutions derived from computational clustering algorithms \cite{zaslavsky2018efficient, regier2015word}. Similarly, research into numeral systems and indefinite pronouns suggests that languages universally optimise their lexicons by balancing semantic precision and cognitive simplicity \cite{kemp2018semantic}.

The linguistic principle of balancing simplicity and informativeness is directly analogous to challenges faced in text clustering. Clustering algorithms aim to partition large datasets into groups, striving for clusters that efficiently summarise data without excessive loss of meaningful detail \cite{van1993categories}. Overly broad clusters may lead to ambiguity and reduced analytical utility, while overly narrow clusters fragment data, complicating interpretation and retrieval. Leveraging linguistic theory, particularly principles from information theory such as Pareto optimality, thus informs clustering strategies by suggesting optimal balances of cluster granularity and coherence. By treating each cluster as a semantic category, analogous to lexical terms in language, we can systematically evaluate clustering solutions for their communicative efficiency.

Furthermore, the coevolution of lexical meaning and pragmatic use highlights the ongoing interaction between learnability and communicative utility in language \cite{brochhagenCoevolutionLexicalMeaning2018}. Simpler lexical forms may initially be preferred due to ease of acquisition, yet pragmatic pressures eventually enrich these forms to express increasingly sophisticated distinctions. As a way to measure how informative a language is, the average cosine similarity between word embeddings of a language can be used \cite{aceves2024human}. Denser conceptual spaces, quantified by higher average cosine similarity between word pairs, align with greater semantic density and efficient communication. We can extend this framework to evaluate clusters in short text data.  Applying this logic to clustering, each cluster's internal semantic cohesion can be assessed using cosine similarity among text embeddings within the cluster. This parallels the approach in language studies but shifts the focus from individual words to clusters of text, treating each bio as a semantic unit. By doing so, we can mimic earlier linguistics methodology to quantify and compare cluster informativeness and coherence, providing a robust measure to balance between cluster density and the meaningfulness of distinctions across clusters. This adaptation supports our overarching aim of identifying an optimal level of abstraction in clustering short texts, resonating with the Goldilocks principle of balancing simplicity and informativeness.

Another way to assess the quality of a cluster is if it can be named \cite{doogan2021topic, newman2010automatic}. A cluster that is able to be named suggests that there is some unifying aspect of the cluster, and that it can be reduced into the low dimensional space of a few words to describe it. However, there are two main concerns with this approach: subjectivity \cite{morstatter2018search}, and usefulness \cite{boyd2014care}. Subjectivity refers to the fact that people may create different names for the same clusters, and since there is no gold standard it is hard to tell if a name is appropriate or not. Usefulness refers to the fact that just because a cluster is able to be named, that does not mean that the cluster is a helpful compression of the data. For instance, consider a dataset consisting of tweets about weather. One cluster contains tweets made by a bot that says ``Check the weather app." While it is straightforward to name this cluster ``Weather app mentions," the cluster doesn't provide deeper insight into the data, apart from indicating the presence of bots.  The tweets containing this reference to an app may have had other distinguishing characteristics, but the simple repetition of this phrase has prompted the clustering and resulting name.  So while the cluster can be named, it fails to offer a richer understanding of its content.

With the advancement of LLMs capable of replicating humans in tasks such as cognitive psychology tests \cite{binz2023using}, market research \cite{li2024frontiers}, and theory of mind tests, LLMs can similarly be used to emulate a human naming a cluster. It is important to note that with the way LLMs work, it is incorrect to describe them as ``coming up" with a name, and more correct to suggest that they are ``stochastic parrots" \cite{bender2021dangers} creating content based on human behaviour. This has been suggested as the cause as to why they perform similarly to humans on these tests: they are taking similar content and parroting it back \cite{binz2023using, li2024frontiers}. This makes it a valid alternative to a human for a tedious task such as naming many clusters.  In our earlier work
we have found that when using LLMs create names, the names given were very similar to human reviewers~\cite{Miller2025}. In addition, despite the stochastic nature of text generation with LLMs the names were also found to have a high degree of internal reliability compared to human reviewers.

This paper aims to establish a methodology for determining the optimal number of clusters by balancing interpretability and complexity. We treat clusters analogously to how languages label semantic categories (e.g. colors) — each cluster name acts as a “word” that compresses and communicates a subset of data \cite{zaslavskyEfficientCompressionColor2018}. Color naming systems across languages evolve towards a Pareto-optimal frontier, balancing between maximizing communicative efficiency and minimizing cognitive complexity. This frontier delineates the set of achievable language systems where no further improvement can be made in informativeness without increasing complexity, and vice versa. Similarly, we believe that clustering will follow a comparable pattern, where the trade-off between cluster granularity and interpretability leads to an optimal frontier that balances informativeness and complexity.

In linguistic terms, we define the environment object $u$ to be a Twitter bio rather than a color chip, and let $m(u)$ represent the meaning distribution over environment objects. In this clustering context, $m$ indexes a particular cluster, and $p(m)$ denotes the probability that a given bio is assigned to cluster $m$. Subsequently, a LLM acts as an  encoder and provides a label (or “word”) $w$ for the cluster, i.e.\ $q(w|m)$, and passes this description along to a LLM that acts as a  decoder that infers which bios $u$ belong to that labeled cluster.

From an information-theoretic perspective, we seek to quantify the trade-off between complexity and accuracy in naming these clusters. We can measure the complexity of an encoding policy $q(w|m)$ using the mutual-information--based metric  \cite{zaslavskyEfficientCompressionColor2018}

\begin{equation} I_q = \sum_{m,w} p(m),q(w|m),\log\frac{q(w|m)}{q(w)}, \label{eq:complexity} \end{equation}

where $p(m)$ is the probability that a bio is assigned to cluster $m$, $q(w|m)$ is the probability of using word $w$ to describe cluster $m$, and $q(w)$ is the marginal probability of using word $w$. Intuitively, this quantity captures how many bits of information must be communicated to distinguish among different clusters—reflecting the “cost” or cognitive load required to maintain interpretability. 

On the other hand, accuracy refers to how well the decoder can recover which bios are being described by a given cluster name: once a word $w$ is communicated, the decoder infers the cluster (and hence the set of possible bios) that the speaker intended. By identifying encoding and decoding strategies that jointly minimize complexity and maximize accuracy (analogues for simplicity and informativeness respectively) we capture the “Goldilocks zone” in clustering, mirroring how color-naming systems in human languages achieve near-optimal compression of semantic categories \cite{zaslavskyEfficientCompressionColor2018}. In our setting, finding this zone ensures that clusters (and their names) are neither too coarse nor too fine, allowing end users to interpret and utilise the resulting groupings in an efficient and meaningful way.

\section{Method}

\subsection{Data Collection}
The data used in this paper consists of 7 datasets collected from Twitter using the Twitter API v2.  Each one was collected using different political keywords and in a different time period, the keyword, time period and number of bios collected is found in Table \ref{tab:bios_keywords}.
\begin{table}[h]
    \centering
    \begin{tabular}{l l r}
        \toprule
        Keywords & Date & Number of Bios \\
        \midrule
        Trump, realDonaldTrump & 04/09/2020 & 38,639 \\
        Biden & 03/11/2021 & 41,927 \\
        Kavanaugh, SCOTUS & 20/09/2018 & 31,028 \\
        realDonaldTrump, Trump & 18/04/2019 & 37,217 \\
        Trump & 23/06/2018 & 21,599 \\
        Trump, realDonaldTrump, POTUS & 18/09/2018 & 32,392 \\
        Election, Midterm & 09/11/2018 & 29,066 \\
        \bottomrule
    \end{tabular}
    \caption{Number of bios mentioning keywords on specific dates}
    \label{tab:bios_keywords}
\end{table}
From these user timelines the username and bio for every user was extracted, however, for this research, only the bio was used. 

\subsection{Clustering}

In this study, clustering methods used by \cite{Miller2025} were also used here. For data preprocessing, emojis were converted to their CLDR Short Name using the Emoji package \cite{2022emoji}. The LLM employed to vectorise the text is the all-MiniLM-L6-v2 model \cite{wang2020minilm, reimers-2019-sentence-bert}.  \cite{reimers-2019-sentence-bert}. For each of the seven datasets, we created clustering models using GMM from 2 clusters up to 50 clusters. A random set was then created to act as a comparison.  This was done by randomly allocating each bio a cluster label, instead of using GMM to create the cluster. `Gemini-pro' was used to create names for each cluster, using the same method from \cite{miller2025moving}.

\subsection{Evaluation Metrics}
\subsubsection{Semantic Density}
For each clustering configuration, from $ K = 2$ to  $K = 50$, each data point was assigned to a cluster. To ensure a balanced representation of different clusters, a stratified sampling approach was used: a random sample of 10,000 data points was drawn while maintaining proportional representation of clusters. To quantify semantic density, pairwise cosine similarities were computed between each sampled data point and a randomly matched point from the same cluster. This ensured that similarity calculations were stratified within clusters. The aggregated mean cosine similarity and the standard error of the mean (SEM) were then computed for each clustering configuration. This process was repeated for all datasets in the study, as well as for a randomly assigned cluster set to serve as a baseline comparison.

\subsection{ Accuracy and Complexity Trade-off}
To calculate Accuracy and Complexity, we randomly sampled 1,000 bios from each clustering model, where a model is defined as a combination of a dataset and a specific number of clusters. Specifically, we applied the Gaussian Mixture Model (GMM) to each of the seven datasets, varying the number of clusters from $ K = 2 $ to $ K = 50 $, resulting in $ 7 \times 49 = 343 $ distinct models. From each model, we selected 1,000 bios, ensuring a consistent sampling approach across all configurations.
We provide Gemini Pro with the following prompt:

``You have a large set of bios from Twitter,

\begin{itemize}
    \item You have a large set of bios from Twitter.
    \item You have clustered them into the following groups: <Cluster Names>
    \item The following bio belongs to only one of these groups. 
    \item Your task is to determine which group the bio belongs to.
    \item You may only choose one group and should respond with only the name of the selected group.
    \item Here is the bio: <sample bio here>.''
\end{itemize}
In this prompt, <cluster Names> refer to the descriptive labels assigned to each cluster by Gemini, and <sample bio> represents an individual Twitter bio that needs to be classified into one of the predefined clusters.

For each answer the LLM provides, we then compare it to the correct name, and class it as correct if its Levenshtein distance is less than 4. The value 4 was chosen because there were instances where the LLM returned the name of the cluster but had slight alterations, such as '**' before or after the name, or slight misspellings of the cluster name. 

To calculate Accuracy, we use the following equation:

$$
\text{Accuracy} = \frac{C}{T}
$$

where:
\begin{itemize}
    \item \( C \): The number of correct cluster assignments made by the LLM.
    \item \( T \): The total number of bios given to the LLM.
\end{itemize}

\begin{equation}
\text{AMI}(U, V) = \frac{\text{MI}(U, V) - \mathbb{E}[\text{MI}(U, V)]}{\text{AVG}[\text{H}(U), \text{H}(V)] - \mathbb{E}[\text{MI}(U, V)]}
\label{eq:AMI}
\end{equation}

To calculate Complexity, we use the Adjusted Mutual Information (AMI) given in (\ref{eq:AMI}) between the actual cluster names and the cluster names produced by the LLM:

$$
\text{Complexity} = \text{AMI}(A, L)
$$

where:
\begin{itemize}
    \item \( \text{AMI} \): Adjusted Mutual Information, a measure of agreement between two clusterings, adjusted for chance.
    \item \( A \): The actual cluster names that each bio belongs to.
    \item \( L \): The cluster names assigned by the LLM.
\end{itemize}

To evaluate the relative performance of GMM clustering against random allocation, we calculated how many standard deviations the average AMI and accuracy values were above the mean of the random baseline. These deviations were used to rank GMM's performance relative to random allocation. Rankings were calculated for each cluster count, allowing us to identify the cluster range where GMM consistently outperformed random allocation in both complexity and interpretability. Error bars in all figures represent the standard deviation across datasets for each cluster count.

We then created a method to try to explain why the LLM was able to correctly identify the cluster names of some bios but not others. We first calculate the cosine similarity between each Twitter bio and the cluster name it came from, and then we calculated the cosine similarity of all other names and found the maximum that  did not correspond to the correct cluster. Cosine similarities were computed to evaluate the alignment between each bio and its associated cluster name, as well as to identify the closest alternative cluster name. We then calculated the difference between the cosine similarity of the cluster name and the alternative cluster name and plotted the distribution of this difference comparing the distributions of when the LLM got it correct and incorrect. To confirm the findings of the visualisation we then used the cosine similarities as predictors in a logistic regression, aiming to predict whether the knowledge of the cosine similarities, the number of clusters in the model, and the data source it came from were enough to predict whether the LLM was able to identify the correct cluster name.

\section{{Results}}
We created cluster models ranging from 2 clusters up to 50 clusters using seven samples of Twitter bios collected around various American political events and figures, including the 2018 midterm election, Brett Kavanaugh’s Supreme Court appointment, Joe Biden, and Donald Trump. Detailed methodology can be found in the Methods section.

\subsection{Semantic Density}

\begin{figure}[ht]
\centering
\includegraphics[width=\linewidth]{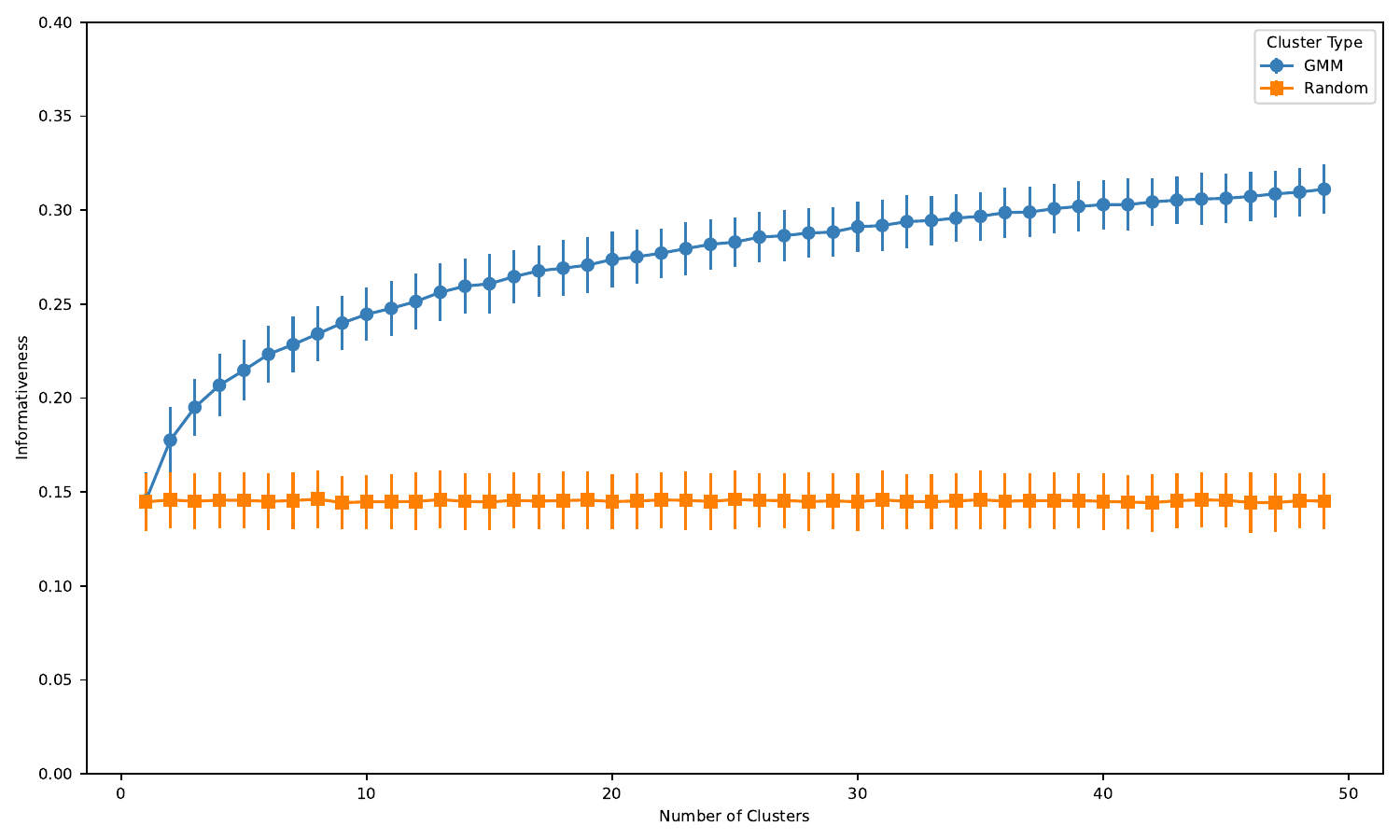}
\caption[Mean Semantic Density]{

Mean semantic density (mean cosine similarity between embeddings of bios within each cluster) as a function of the number of clusters (2–50) for Gaussian Mixture Model (GMM) clustering and randomly assigned clusters. Data points represent averages calculated from up to 10,000 random pairs per cluster, capped at the maximum number of available pairs. Error bars indicate the standard error of the mean semantic density across clusters. GMM clustering results are shown with circles and solid lines, while randomly assigned clusters are shown with squares and dashed lines. The y-axis is capped at 0.4 to focus on the observed range of semantic density.}
\label{fig:Informativeness}
\end{figure}

Figure \ref{fig:Informativeness} illustrates how semantic density (as measured by mean cosine similarity) changes with the number of clusters for both GMM-based clustering and randomly assigned clusters. As the number of clusters increases, GMM clustering increases in semantic density, while the random assignment stays at 0.15, indicating that GMM is capturing more meaningful groupings in the data. The standard error of the mean across datasets for both GMM and random clusters is quite low, showing that there is not much deviation in the datasets.
The rise in semantic density for the GMM assignment begins to level off around the mid-range of cluster counts, highlighting a Goldilocks trade-off: using too few clusters may obscure key distinctions among bios, whereas using too many can fragment the data and complicate interpretation.

\subsection{Information Metric}

\begin{figure}[ht]
\centering
\includegraphics[width=\linewidth]{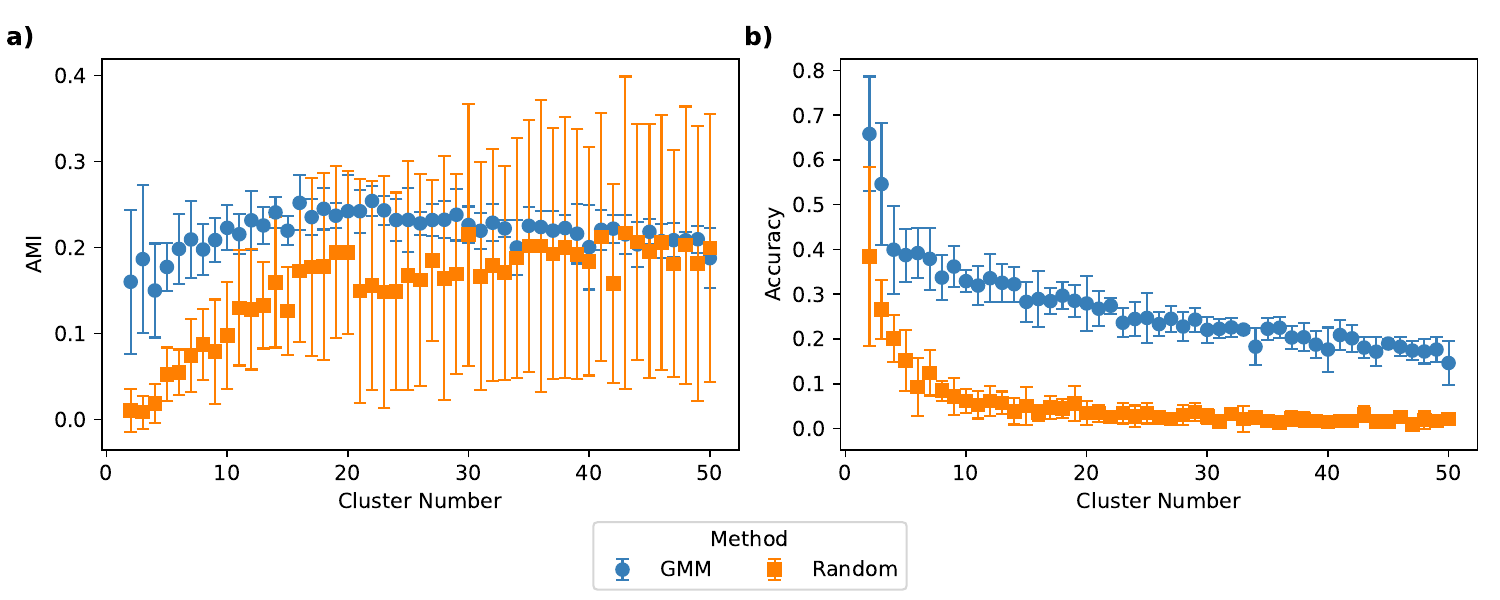}
\caption[AMI and Accuracy of LLM ability to recognise clusters through the names of the clusters.]{

Comparison of Adjusted Mutual Information (AMI) and Accuracy between Gaussian Mixture Model (GMM) clustering and a random baseline. A LLM was given 1000 random bios at each clustering level and asked to identify which cluster it belonged to having been given the cluster names. (a) AMI quantifies the agreement between true cluster assignments and predicted cluster assignments, adjusted for chance. Higher values indicate better clustering performance. (b) Accuracy measures the proportion of bios that were correctly assigned to a cluster when given only the cluster names, representing the interpretability of the clustering process. Each data point represents the mean value across experiments for a given number of clusters, with error bars indicating the standard deviation. GMM consistently outperforms the random baseline in both AMI and Simplicity at lower levels but at high number of clusters, the AMI between GMM and Random is almost the same.}
\label{fig:Simplicity}
\end{figure}

Figure \ref{fig:Simplicity} compares the complexity (Adjusted Mutual Information, AMI) and interpretability (accuracy) of clusters generated using Gaussian Mixture Models (GMM) and random allocation. AMI increases with the number of clusters for both methods, reflecting a higher agreement between true and predicted cluster assignments. However, while GMM clustering initially outperforms random allocation, its AMI peaks around 20 clusters before gradually declining. In contrast, the random allocation shows a continued increase in AMI, resulting in comparable or even higher values at higher cluster numbers. The higher standard deviation observed in random allocations suggests that outliers might be inflating the average AMI.

For accuracy, both methods show a consistent decline as the number of clusters increases. This trend aligns with expectations, as more clusters introduce greater uncertainty and more chances for incorrect assignments. The standard deviation for accuracy is relatively low across datasets, particularly at higher cluster counts, indicating consistent performance. To determine the optimal number of clusters, it is essential to balance AMI and accuracy, aiming to maximize both metrics while accounting for their respective trade-offs.

\begin{figure}[ht]
\centering
\includegraphics[width=\linewidth]{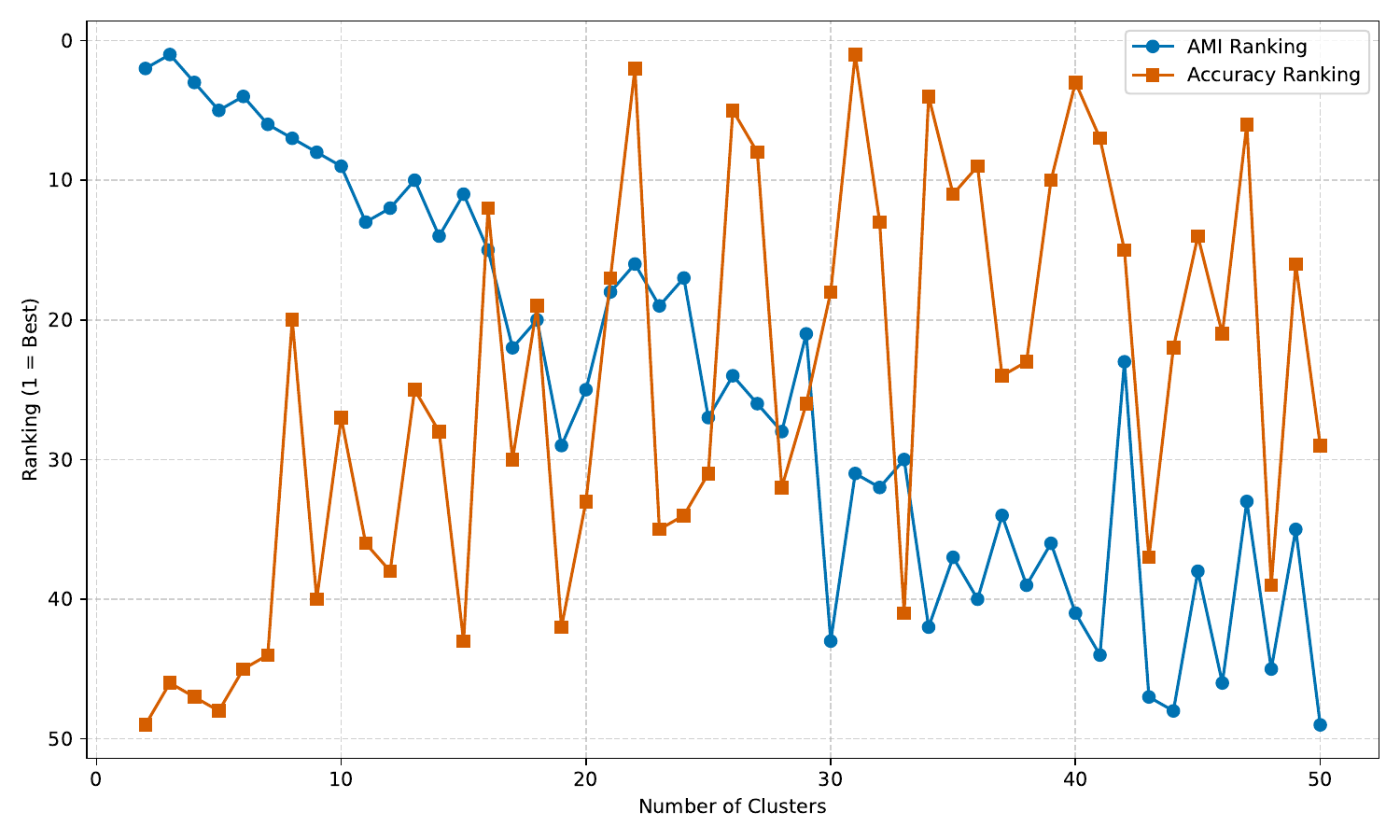}
\caption[Ranking of Clusters compared to Random assignment]{

Comparison of GMM-based clustering versus randomly assigned clusters, ranked by how many standard deviations each method’s performance (AMI and Simplicity) deviates from Random. For each number of clusters, we compute the difference between GMM and Random in standard deviation units, then assign a rank (1 = best). Higher rankings indicate that GMM’s performance is consistently farther above Random, whereas lower suggest Random rivaling or exceeding GMM. By visually identifying where rankings are highest for both AMI (blue) and Simplicity (red), one can approximate the optimal number of clusters that best balances complexity and interpretability.} \label{fig:ranking_vs_clusters}
\end{figure}


To identify a value that simultaneously maximises AMI and accuracy, we calculate how many standard deviations the average value is above the mean of random allocations as shown in Figure \ref{fig:ranking_vs_clusters}, and then rank each value of $K$ based on how many standard deviations the mean is above the random mean. Rankings are used instead of absolute values to mitigate the influence of extreme deviations, as absolute values can disproportionately skew interpretations and obscure meaningful comparisons across clusters.

As a general trend, the AMI ranking increases with the number of clusters, reflecting increasing complexity. Conversely, the accuracy ranking decreases as the number of clusters increases, indicating that the language model (LLM) becomes more accurate than random assignments at higher cluster counts. Accuracy exhibits greater fluctuations than AMI, with an overall trend of improvement interspersed with periods of decline in rankings. Notably, the two lines intersect multiple times around cluster values 16–22, suggesting that this region could represent a ``Goldilocks zone" for clustering. While there are a few intersections at higher cluster counts, they appear less consistent than those observed in this range.

\subsection{Cosine Difference as an Explanation of Accuracy}
\begin{figure}[ht]
\centering
\includegraphics[width=\linewidth]{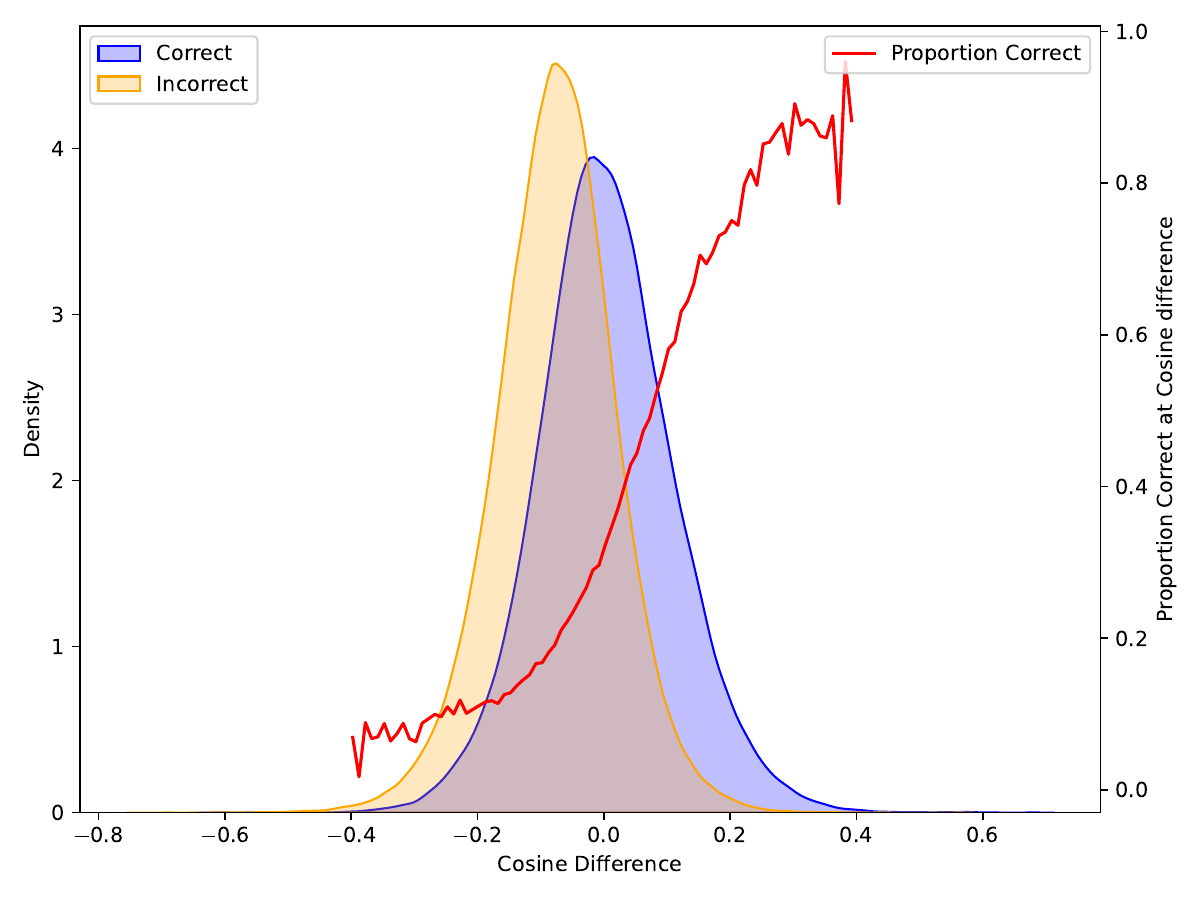}
\caption[Cosine distance as an explanation of model accuracy]{

This figure illustrates the kernel density estimate (KDE) distributions of cosine similarity differences for bios. The cosine similarity difference is calculated as the difference between the similarity of a bio with its assigned cluster name and the similarity of the same bio with the most similar incorrect cluster name. The blue curve represents the density of bios correctly identified by the LLM, while the orange curve corresponds to bios incorrectly identified. The red line, plotted on a secondary y-axis, shows the proportion of correct cluster name assignments across binned cosine similarity differences (0.01 increments) within the range -0.4 to 0.4, as data outside this range was sparse.} \label{fig:cosine_difference}
\end{figure}

The previous section demonstrated that Gemini Pro's ability to correctly assign bios to clusters is influenced by the semantic relationships between bios and cluster names. However, this does not fully explain why the model is more successful for some bios than others. To investigate this, we analyse the role of cosine similarity between bios and cluster names as a potential predictor of correct classification.

A reasonable assumption is that Gemini Pro will be more likely to correctly classify a bio when its semantic similarity to the correct cluster name is significantly higher than its similarity to any incorrect cluster name. If the correct cluster name is much closer in embedding space than any alternative, the model is expected to assign the bio correctly. Conversely, if an incorrect cluster name is nearly as similar as the correct one, misclassification becomes more likely.

To test this hypothesis, we compute the cosine similarity difference for each bio:

\begin{equation}
    Cosine Difference = \cos(B, C_{\text{correct}}) - \max_{C_{\text{incorrect}}} \cos(B, C_{\text{incorrect}})
\end{equation}

where:
\begin{itemize}
    \item $\cos$ is the cosine similarity between two embeddings
    \item \( B \) represents the bio embedding,
    \item \( C_{\text{correct}} \) is the embedding of the correct cluster name,
    \item \( C_{\text{incorrect}} \) represents all other cluster names except the correct one.
\end{itemize}

This difference captures the semantic separation between the correct cluster name and the closest incorrect alternative. Higher values indicate that the correct name is distinctly more similar to the bio than any competing name, whereas lower values suggest potential ambiguity.

Figure~\ref{fig:cosine_difference} presents the distribution of cosine similarity differences for correctly and incorrectly classified bios. The blue density curve represents bios that were correctly assigned, while the orange density curve represents those that were misclassified. 

On a secondary y-axis (in red), the figure also shows the proportion of correct assignments as a function of the cosine difference. As expected, a positive cosine difference (where the bio is closer to the correct name than any other) correlates with higher classification accuracy. The probability of correct classification surpasses 50\% when the cosine difference exceeds 0.1, and at 0.2, the model correctly assigns the bio approximately 75\% of the time.

\subsection{Logistic Regression Results}
To further validate this trend, we apply logistic regression, using cosine similarity differences, the number of clusters in the model, and the dataset source as predictors of whether Gemini Pro correctly assigns a bio to its cluster. This regression allows us to quantify how much cosine similarity contributes to classification accuracy, controlling for other potential factors.
Given from previous results, we know that cluster number would most likely be a significant predictor, we include that too, as well as the dataset which may impact it, i.e. the LLM may struggle with one group of bios over another. 

\begin{table}[ht!]
\centering
\begin{tabular}{lrrrr}
\hline
\textbf{Variable} & \textbf{Coef.} & \textbf{Std. Err.} & \textbf{z-value} & \textbf{p-value} \\
\hline
\texttt{Intercept} & -0.8951 & 0.028 & -32.416 & 0.000 \\
\texttt{Correct Cosine Sim.} &  8.1991 & 0.091 & 89.857 & 0.000 \\
\texttt{Incorrect Cosine Sim.} & -4.9648 & 0.097 & -50.977 & 0.000 \\
\texttt{Interaction Term} & -2.6215 & 0.266 & -9.855 & 0.000 \\
\texttt{Clusters Count} & -0.0276 & 0.000 & -83.584 & 0.000 \\
\texttt{Filename: Kavanaugh} & -0.0074 & 0.016 & -0.466 & 0.641 \\
\texttt{Filename: Trump (06/2018)} &  0.0640 & 0.016 & 4.096 & 0.000 \\
\texttt{Filename: Trump (09/2020)} &  0.0061 & 0.016 & 0.392 & 0.695 \\
\texttt{Filename: Trump (09/2018)} &  0.0008 & 0.016 & 0.053 & 0.958 \\
\texttt{Filename: Midterm (11/2018)} &  0.0918 & 0.016 & 5.913 & 0.000 \\
\texttt{Filename: Trump (04/2019)} &  0.1183 & 0.016 & 7.619 & 0.000 \\
\hline
\end{tabular}
\caption{Logistic regression results predicting whether a bio is correctly assigned based on cosine similarities, the number of clusters, and filename controls.}
\label{tab:logistic_regression}
\end{table}

Table~\ref{tab:logistic_regression} presents the coefficients (\(\beta\)), standard errors, and statistical significance of the logistic regression model. 

The intercept (\(\beta = -0.8951\)) represents the log-odds of correct classification when all predictors are set to zero. Converting this log-odds value into a probability gives:

\[
\text{Probability} = \frac{e^{-0.8951}}{1 + e^{-0.8951}} \approx 0.29 \quad (29\%).
\]

This represents the baseline probability of correct classification when both cosine similarities are zero, and the number of clusters is at its minimum (2 clusters).

An increase in the cosine similarity to the correct cluster name (\(\beta = 8.1991, p < 0.001\)) substantially raises the odds of correct classification. For example, increasing \texttt{Correct Cosine Sim.} by 0.1 while holding all other predictors constant changes the log-odds to:

\[
\text{Log-Odds} = -0.8951 + (8.1991 \times 0.1) = -0.8951 + 0.8199 = -0.0752.
\]

Converting this to probability:

\[
\text{Probability} = \frac{e^{-0.0752}}{1 + e^{-0.0752}} \approx 0.48 \quad (48\%).
\]

Thus, a 0.1 increase in cosine similarity to the correct cluster name increases the probability of correct classification from 29\% to 48\%.

Similarly, an increase in the cosine similarity between a bio and an incorrect cluster name (\(\beta = -4.9648, p < 0.001\)) decreases the odds of correct classification. If we set the cosine similarity between a bio and the correct cluster name to zero and increase \texttt{Incorrect Cosine Sim.} by 0.1, the log-odds change as follows:

\[
\text{Log-Odds} = -0.8951 + (-4.9648 \times 0.1) = -0.8951 - 0.4965 = -1.3916.
\]

Converting this to a probability:

\[
\text{Probability} = \frac{e^{-1.3916}}{1 + e^{-1.3916}} \approx 0.20 \quad (20\%).
\]

Thus, a 0.1 increase in cosine similarity to an incorrect cluster reduces the probability of correct classification from 29\% to 20\%.

If both cosine similarities are set to 0.1, the log-odds include their interaction term (\(\beta=-2.6215\)):
\[
\text{Log-Odds} = -0.8951 + (8.1991 \times 0.1) + (-4.9648 \times 0.1) + (-2.6215 \times 0.1 \times 0.1).
\]
Substituting the values:
\[
\text{Log-Odds} = -0.8951 + 0.8199 - 0.4965 - 0.0262 = -0.5980.
\]
Converting this to a probability:
\[
\text{Probability} = \frac{e^{-0.5980}}{1 + e^{-0.5980}} \approx 0.35 \; (35\%).
\]

This suggests that there is a 35\% chance of the LLM being able to correctly identify the cluster the bio belongs to, if both the correct cluster name and the closest incorrect cluster name both have a cosine similarity of 0.1.

Of the seven datasets included in this analysis, there were statistically significant differences in the model’s predictive performance, indicating that some datasets were more conducive to creating informative cluster names than others. Although these differences reached statistical significance, the associated effect sizes (i.e., the corresponding $\beta$ values) were relatively small, suggesting modest practical impact. In contrast, the total number of clusters was a robust predictor, such that higher levels of clustering complexity had a more pronounced detrimental effect on correct cluster assignment than did a bio having elevated similarity to an incorrect cluster name. Consequently, while dataset variations are detectable and significant, the influence of cluster count appears to be more critical for the interpretability and reliability of these model-derived cluster labels. 

\section{Discussion}

These results show that a key factor in whether a LLM can correctly name a cluster is due to how semantically similar the cluster name is to an example from the cluster. However, if other names are too semantically similar this will lead to incorrect results. This suggests that a potential way of evaluating a cluster name can be down to how semantically similar it is to the points within the cluster, and how disimilar those points are to other cluster names.

The outcomes of this paper align  with earlier work that highlights the delicate balance between simplicity and informativeness in both linguistic categorisation and data clustering \cite{aceves2024human, carr2020simplicity, zaslavskyEfficientCompressionColor2018}, showing that it is possible to balance both simplicity and informativeness. Recent scholarship has demonstrated how human languages evolve toward minimal yet sufficiently expressive sets of lexical categories, aligning with the principle that communication systems must not overburden their users with superfluous distinctions. Analogously, our clustering approach indicates that there are sets of clusters able to balance being simple enough to easily understand, while also communicating the core semantic content of Twitter bios. This parallels colour naming systems, where near-optimal compression ensures that speakers can communicate crucial perceptual distinctions without proliferating excessive category boundaries \cite{zaslavskyEfficientCompressionColor2018}.

Our results further reveal how interpretable cluster labels, generated by large language models (LLMs), can facilitate enhanced understanding of social media data while remaining sufficiently straightforward for end-users. Specifically, as the number of clusters increases, the interpretability of the resulting labels declines, in keeping with the notion that greater granularity fragmenting the data to a point of diminishing returns. This finding is essential for future research in clustering, suggesting that the “Goldilocks zone” of clustering not only underscores how best to delineate semantic categories, but also how to optimally communicate these categories to non-expert audiences. By striking a balance between preserving essential distinctions and avoiding overwhelming the user with excessive detail, research efforts can better tailor cluster counts and naming strategies to suit practical needs in clustering of short-text.

Moving forward, several considerations warrant attention. First, although LLM-generated names exhibited reliable performance in our study,there is a large stochastic component involved,especially in using LLMs through APIs where there is no way to ensure replicability in the results. For more reliability it is reccommended to use LLMs locally loaded so that one can ensure the same results every time. Future work could incorporate human judgments, comparing them with LLMs, and testing what other factors may affect the ability to correctly identify clusters, and if humans are able to better identify them than LLMs. Second, the study’s focus on Twitter bios may limit generalisability, signalling the value of replicating the research on other data domains to validate these findings. Finally, while the use of semantic density measures and logistic regression provides robust quantitative insights, further refinement—such as the incorporation of alternative embedding methods may lead to better results. 

\section{Conclusion}
This study explored the delicate balance between simplicity and informativeness in clustering short text data, framed through the lens of linguistic theory and the concept of a 'Goldilocks zone.' Three key experiments were conducted:

Firstly, semantic density was measured to assess how effectively Gaussian Mixture Model (GMM) clustering captured meaningful groupings compared to random allocation. Higher semantic density was observed GMM as the cluster resolution increased, however, there was a steady drop-off in the gains that increasing the clustering resolution achieved, suggesting that there is non-linear relationship between informativeness and the clustering resolution.

Secondly, cluster interpretability and complexity were evaluated using Adjusted Mutual Information (AMI) and accuracy metrics, with cluster labels generated by LLMs, and clusters generated by both GMM and random allocation. Comparing the GMM and random clusters revealed a clear trade-off: the AMI of GMM clusters versus random was initially much higher, but declined as the number of clusters increased, conversly the opposite occurred with asccuracy. This demonstrates the trade-off between informativeness and complexity in clustering, and but also showed a period between 16-22 where the GMM clusters had a high AMI and accuracy relative to the random model, demonstrating a potential Golidlocks zone at this section.

Finally, logistic regression analysis examined factors influencing the accuracy of cluster identification by the LLM. Results demonstrated that semantic similarity between bios and corresponding cluster names strongly predicts successful identification, suggesting effective cluster naming requires distinct, semantically coherent labels.

Collectively, these experiments provide practical guidelines for optimising cluster numbers and underline the importance of carefully designed, semantically clear cluster labels.

Future research should replicate these methods across varied datasets and further investigate alternative embedding models. Practically, calibrating cluster counts carefully and deploying locally hosted LLMs can enhance reliability and interpretability, aligning clustering methods closely with principles of efficient human communication.


\bibliography{Main_File}
\end{document}